# Conditional Activation for Diverse Neurons in Heterogeneous Networks

Albert Lee[1], Bonnie Lam[1], Wenyuan Li[1], Hochul Lee[1], Wei-Hao Chen[2], Meng-Fan Chang[2], and

Kang. -L. Wang[1]

[1] University of California, Los Angeles, [2]National Tsing-Hua University

**Abstract**

In this paper, we propose a new scheme for modelling the diverse behavior of neurons. We introduce the "conditional activation", in which a neuron's activation function is dynamically modified by a control signal. We apply this method to recreate behavior of special neurons existing in the human auditory and visual system. A heterogeneous multilayered perceptron (MLP) incorporating the developed models demonstrates simultaneous improvement in learning speed and performance across a various number of hidden units and layers, compared to a homogeneous network composed of the conventional neuron model. For similar performance, the proposed model lowers the memory for storing network parameters significantly.

## I. Introduction

Machine learning (ML) methods have recently demonstrated outstanding results with structures such as recurrent neural networks (RNNs), feedforward neural networks (FNNs), and convolutional neural networks (CNNs), including, the use of numerous methods such as drop-out/drop-connect [1][2], batch normalization [3], annealed/momentum learning rates [4], and data augmentation schemes [5] to allow for better generalization of learning result. These schemes have enabled state-of-the-art results in recognition accuracy on many datasets, at the cost of slower convergence, larger network size, or increased amount of training data. However, despite these innovations, an identical neuron model is used throughout the entire artificial neural network. This homogeneity in the neurons is thus a limiting factor in further advancement in performance, as a homogeneous system is by far less versatile and adaptive than its heterogeneous counterpart, as proven in numerous fields including communications [6], evolution/society [7], and neural networks [8]. Furthermore, the homogeneity of current AI systems neither allows the capability to contain intrinsic functions to drive the system toward a goal, nor rules that guide and govern the learning process within safety bounds.

In contrast, the biological neural system contains a rich variety of neurons varying in structures, electrical properties, synaptic connections, as well as spatial and timing response [9]. Different neuron types result in high versatility and complementary computing capabilities, and their hard-wired anatomy further compose the intrinsic functions, rules, and values in the neural system. (For example: functions existing at birth which are not obtained or changed through learning, such as the auditory and visual systems [10]; non-learnable rules that govern behavior and safety, such as reflexes [11]; unconditioned reward values that motivate behavior, learning, and pleasure, such as enjoying music [12]–[14]). Thus, in order to further advance artificial intelligence, it is essential to understand and to utilize neural diversity and their anatomy for next generation AI.

In this work, we present a new scheme for modelling the diverse response of neurons that are critical in human speech and image processing, and exhibit significantly different dynamic behavior than the traditional integrate-and-fire neuron. The remainder of this paper is organized as follows: Section II introduces the proposed conditional activation scheme. Section III demonstrates the application of the conditional activation in creating diverse neuron models. Section IV shows learning results on a heterogeneous network composed of the developed models for the MNIST dataset. Section V summarizes this paper with discussions on future research directions.

## II. The Conditional Activation Scheme

### II.A. Modulation of a Neuron's response

In the neural system, the electrical response of a neuron can be dynamically modulated through several mechanisms. Chemical messages such as neurotransmitters and neuromodulators target receptors on a neuron leading to changes in various features [15][16]. Neurotransmitters are released at synapses affecting a small, local region, while neuromodulators are diffused into the intracellular neural tissue, thereby affecting a larger cluster of neurons. Activation or deactivation of genetic expressions through transcription factors may also change a neuron's functions [17].

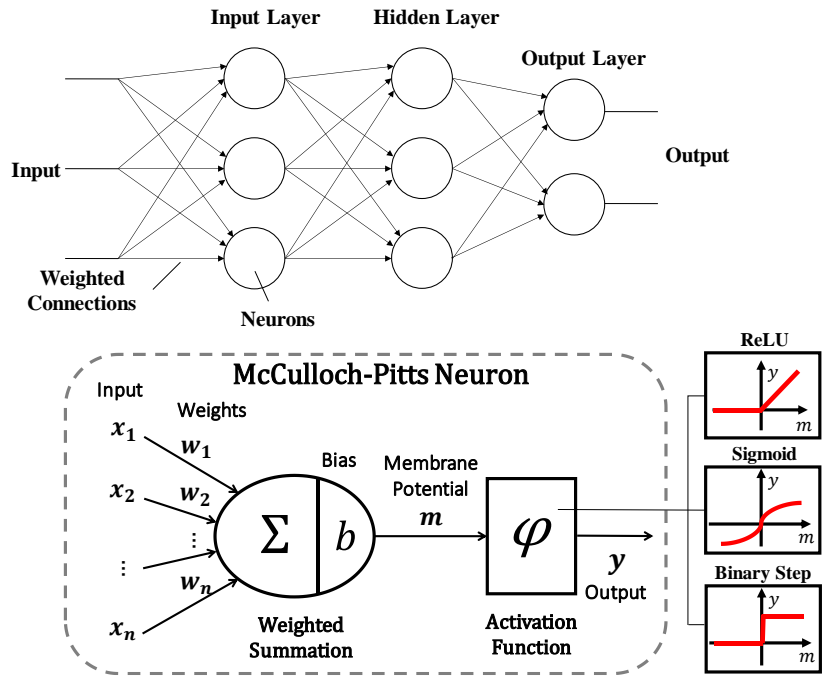

Fig.1 Structure of a multi-layer perceptron (MLP) incorporating the McCulloch-Pitts neuron model, which computes the weighted sum of its inputs passed through an activation function.

### II.B. Formulation of the conditional activation scheme

A common context of the mechanisms in Sec. II.A is that a control signal changes the neurons response. The conditional activation scheme models this phenomenon by allowing a control signal to change a neuron model's activation function. Fig.1 shows the conventional McCulloch-Pitts neuron employed in machine learning applications [18]. In this case, the neuron primarily composes of two parts: a weighted sum and an activation function. During forward propagation, the membrane potential $m$ is computed as the weighted sum ($\sum x_i w_i$) plus a bias $b$. The result is then passed through the activation function, often a nonlinear thresholding function such as the rectified linear unit (ReLU), sigmoid, or step function. Formally,

$$y = \varphi(m) = \varphi\left(\sum_{i=1}^{n} x_i w_i + b\right)$$

where $x_i$ is the neuron's input, $w_i$ is the connection weights, $\varphi$ is the activation function, and $y$ is the output of the neuron.

The conditional activation scheme is shown in Fig.2. In this scheme, the activation has a set of response functions $\varphi_1 \ldots \varphi_k$ and takes in a set of input control channels $L = [L_1 \ldots L_n]$. Based on the pattern of the controls, one of the response functions is selected as the activation function of the neuron. Formally,

$$\varphi_i \in [\varphi_1(m), \varphi_2(m) \ldots]$$

$$y = \varphi_i(m)$$

where $i = f_{cond}(L) \in \mathbb{N}$ is the mapping from $L$ to each response function.

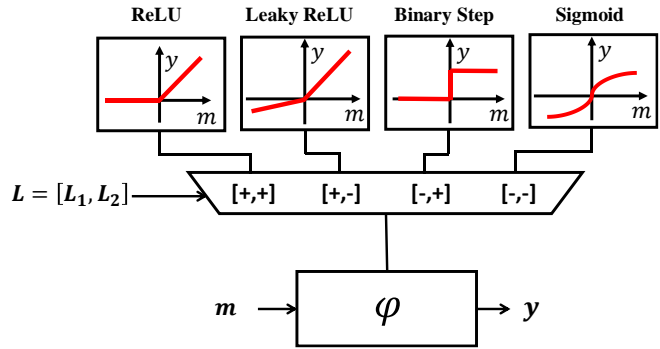

Fig.2 The conditional activation scheme. Here, the controls $L$ composes two channels $L_1$ and $L_2$, and the response functions $\varphi_i$ contains the ReLU, Leaky ReLU, Binary step, and Sigmoid functions. When both control channels are active, the activation function becomes a ReLU. When channel $L_1$ is active and $L_2$ is inactive, a leaky ReLU is selected. When channel $L_1$ is inactive and $L_2$ is active, a step function is selected. When both channels are inactive, the sigmoid function is selected.

### III. Conditional Activation in modelling diverse neuron functions

#### II.A. Coincidence/synchrony detection and max functions

Cells in the visual and auditory cortex show a variety of neuron functions. The octopus cells in the cochlear nucleus detects synchronous activity, firing only when multiple input synapses are simultaneously active [19]. It does not fire when two inputs arrive with a time delay, regardless of the amplitude and frequency of the signals. The ability to detect synchrony plays a vital role in human interpretation of temporal characteristics [20]–[22]. In fact, clinical studies have shown that individuals with damaged synchrony detection causes difficulty in speech interpretation [19]. The V4 neuron (also known as complex cell [23]) in the visual cortex computes the maximum value among its input activities [24]. It determines the region of focus in the field of view and provides attention modulation [25], and creates invariance, the ability to provide robust recognition despite shifting, resizing, and other transformations in the input. The max pooling layer widely adopted in RNNs is an extreme abstraction of this neuron's function.

#### II.B. synchronous and max neuron models

To model the activities of these neurons, we first construct the truth table of their electrical response with two inputs (In1 and In2). We then map the inputs to $L$ and $m$, respectively, and translate firing to "above threshold" and silent to "below threshold", as shown in Fig.3. Here, we set the threshold to 0. Inspecting the translated table, we see that for synchronous detection, the output is a constant 0 when the control $L$ is negative (inactive). However, when $L$ is positive (active), the output is $m$ when $m$ is positive and 0 when $m$ is negative, characterized by the ReLU function. This gives us the conditional activation for the synchronous detection:

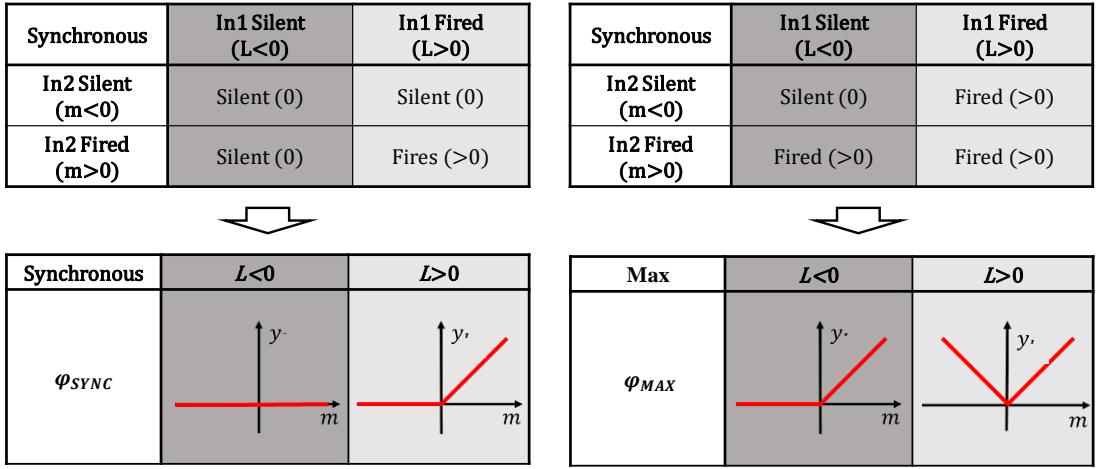

Fig.3 Using the conditional activation to model the synchronous detection and max functions. The truth table of the neuron's activity is mapped to the activation's inputs, then the mathematical representation is constructed.

$$\varphi_{SYNC,i} \in [0, ReLU];$$

$$i = f_{cond}(L) = \begin{cases} 1 & L < 0 \\ 2 & L > 0 \end{cases}$$

For the max function, when $L$ is negative, the output is again characterized by the ReLU function. However, when $L$ is positive, the output is equal to $m$ when $m$ is positive and the inverse of $m$ when $m$ is negative, characterized by the absolute function. This gives us the CA for the max function:

| Parameters | | $L > 0$ | | $L < 0$ | |
|---|---|---|---|---|---|
| | | $m > 0$ | $m < 0$ | $m > 0$ | $m < 0$ |
| $\varphi_{SYNC}$ | $\partial y/\partial m$ | 1 | 0 | 0 | 0 |
| | $\partial y/\partial L$ | 0 | 0 | 1 | 0 |
| $\varphi_{MAX}$ | $\partial y/\partial m$ | 1 | -1 | 1 | 0 |
| | $\partial y/\partial L$ | 0 | 1 | 0 | 0 |

Table 1. Back propagation parameters of $\varphi_{SYNC}$ and $\varphi_{MAX}$ used in this experiment. The back-propagation on the control $L$ is 1 when the neuron's output differs from the output generated by the ReLU function.

$$\varphi_{MAX,i} \in [ReLU, abs];$$

$$i = \begin{cases} 1 & L < 0 \\ 2 & L > 0 \end{cases}$$

The back-propagation parameters of the output with respect to $m$ ($\partial y/\partial m$) is modified according to the selected response function. For the back propagation on $L$ ($\partial y/\partial L$), we simply set it as 1 when the control caused the neuron's response to differ from its original response. In the case of synchronous detection, this occurs when $L$ is negative and $m$ is positive. In the case of the max function, this occurs when $L$ is positive and $m$ is negative.

## IV. Experimental Results

We evaluate our model on MNIST, a hand-written digit dataset containing 60,000 training images and 10,000 testing images divided among 10 classes [26] (digits 0-9). Each grayscale image is a size of 28×28 pixels normalized to the range [0,1]. Two MLP networks were constructed: the reference homogeneous network composed entirely of conventional models with the RELU activation, and the proposed heterogeneous network with the second hidden layer composed of 50% synchronous conditional activations and 50% max conditional activations, as shown in Fig.4. The *L* of each neuron is connected to the *m* of a random neuron in the same layer. The output of the last layer is fed into a softmax classifier. Training was done in batch sizes of 100 and weights were updated using the Adam [27] gradient descent algorithm.

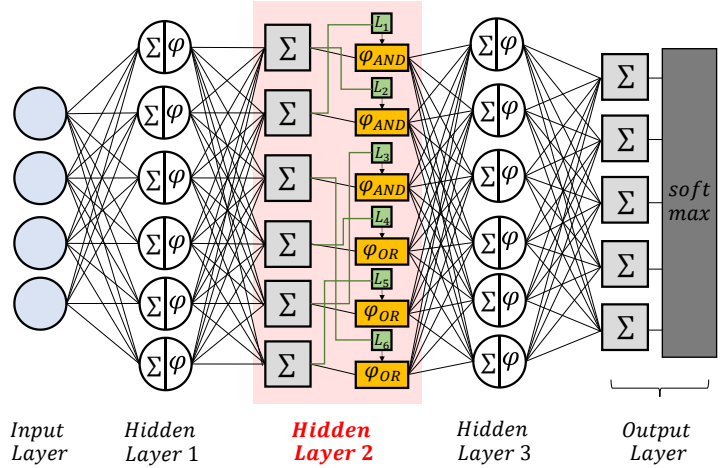

Fig.4 MLP network structure for evaluating the proposed neuron model

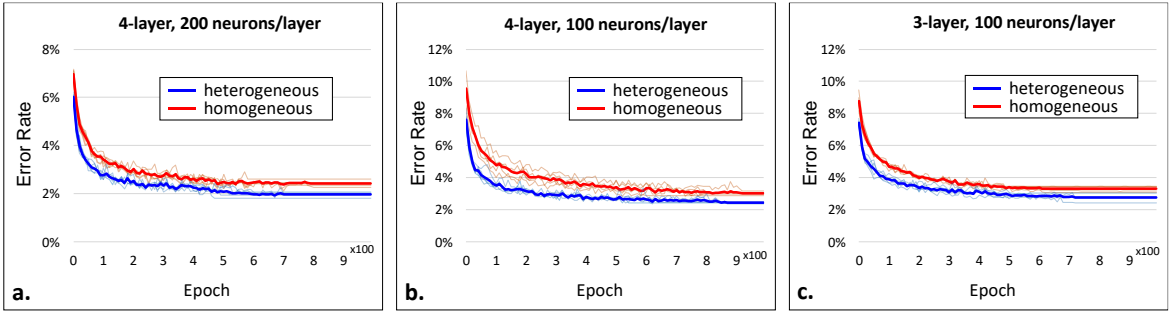

Fig.5 Recognition error as a function of training epochs for (a) a 4-layer, 200-neuron per layer MLP. (b) a 4-layer, 100-neuron per layer MLP, and (c) a 3-layer, 100-neuron per layer MLP

Fig.5(a) shows the recognition error as a function of epochs for a 4-layer MLP (3 hidden, one output) with 200 neurons per input/hidden layer. The final recognition error (i.e. when training set overfits) averaged over 5 trials of the homogeneous conventional model is 2.42%, while that of the inhomogenseous network incorporating the proposed model achieves 1.96%, demonstrating a ~20% improvement in error rate. The number of epochs for the former to reach 99% of the final error rate is 110 epochs, while that of the proposed takes only 70 epochs, a speedup of 36%.

To generalize the improvement of the proposed model compared to the conventional model, we reduce the number of neurons in each layer from 200 to 100. Expectedly, the error rate increases as the number of neurons per layer is reduced, as shown in Fig.5(b). The proposed models achieve an improvement of 20% and 46% in the error rate and training speed, respectively. We also look at the effect of reducing the number of hidden layers in the network. Our proposed model achieves improvement in error rate and training speed of 17% and 28%, at 3-layer, 100 neurons per layer.

The advantages of the heterogeneous network is not only restricted to enhancements in recognition rate and training speed, but also extends to memory capacity. Table 2 shows a summary of the recognition error rate for different network structures, along with the memory needed to store the network. The necessary memory to store weights from a layer of $n_1$ neurons to $n_2$ neurons is $n_1 n_2$, and the memory for storing biases for $n_1$ neurons is $n_1$. For the proposed network, an additional $n_1$ parameters are needed to store the $L$ input index of each cell. The proposed scheme outperforms a conventional 4-layer, 400 neurons per layer network with a modified 4-layer, 200 neurons per layer network, with a >60% reduction in the memory storage while simultaneously improving error rate and training time.

| Network Structure | | Parameters | Error Rate | Training Time* |
|---|---|---|---|---|
| 4-layer 400 neurons | Homogeneous | 638,810 | 2.05% | 70 |
| | Heterogeneous | 639,210 | 1.79% | 50 |
| 4-layer 200 neurons | Homogeneous | 239,410 | 2.42% | 110 |
| | Heterogeneous | 239,610 | 1.96% | 70 |
| 4-layer 100 neurons | Homogeneous | 99,710 | 3.03% | 260 |
| | Heterogeneous | 99,810 | 2.43% | 140 |
| 3-layer 200 neurons | Homogeneous | 199,210 | 2.47% | 100 |
| | Heterogeneous | 199,410 | 2.15% | 80 |
| 3-layer 100 neurons | Homogeneous | 89,610 | 3.31% | 180 |
| | Heterogeneous | 89,710 | 2.75% | 130 |

*Epochs to 99% final recognition error rate

Table 2. Recognition error rate, training time, and number of network parameters for different network structures.

## V. Summary and Future Directions

We have introduced a novel scheme for modelling the diverse neural activity in the nervous system, particularly the dynamical modulation of a neurons response. Using this method, we recreated the activity of two special neuronal functions, the synchronous detection and the max function. Preliminary experiment results show that the heterogeneous network incorporating these models show improved recognition rate as well as training epochs across various network configurations, without the penalty in convergence speed, network size, and training data as previous machine learning algorithms.

The research directions presented in this paper are intended to further advance machine learning performance through neural diversity. In addition to combining the present proposed model with other machine learning algorithms (e.g. Dropout, DropConnect) and network structures (RNN, spiking,…), the following research directions are also of great interest: (i) Further increasing the diversity in an inhomogeneous neural network, by exploring the different application-specific neurons existing in the human nervous system that are known to be critical to system function, and develop models that represent their behavior at a variety of abstractions. For example, the Nobel Prize winning discovery of place cells and grid cells [28] may prove essential to navigation tasks. (ii) Understanding how learning is affected by neuromodulation, as there exist neurons that are very plastic and also cells that do not learn at all. Furthermore, the learning is also changed dynamically by the same mechanisms in Sec. II. (iii) Exploring the organization of neural functions in both artificial and biological neural networks. The hard-wired anatomy of neuron functions create intrinsic functions, rules, and values that govern our behavior and learning. We thus need to explore the organization of neuronal functions in both biological and the artificial networks, to create a methodology for building intrinsic functions in the system that lead to goal-driven unsupervised learning. At the same time, these functions will compose rules in the operation and growth of the network, such that the network cannot outstep certain boundaries. Interestingly, the mapped synchronous detection and the max functions are identical to the AND and OR functions in computing logic, which may provide a guideline for this purpose. The combined results from these three research directions will allow AI systems to be smaller, faster, and more powerful; at the same time able to achieve tasks without the need for large amounts of training data, while ensuring safety in operation and learning.

**Acknowledgements**